\documentclass[10pt,twocolumn,letterpaper]{article}

\usepackage{iccv}
\usepackage{times}
\usepackage{epsfig}
\usepackage{graphicx}
\usepackage{amsmath}
\usepackage{amssymb}
\usepackage{kotex}
\usepackage{multirow}
\usepackage{booktabs}
\usepackage{subfigure}
\usepackage[titletoc,title]{appendix}

\usepackage[pagebackref=true,breaklinks=true,letterpaper=true,colorlinks,bookmarks=false]{hyperref}

\iccvfinalcopy 

\def\iccvPaperID{****} 

\ificcvfinal\fi
\begin{document}

\title{EXTD: Extremely Tiny Face Detector via Iterative Filter Reuse}

\author{
YoungJoon Yoo\thanks{Clova AI Research, NAVER Corp. We follow the alphabetical order except the first author.}\\
{\tt\small youngjoon.yoo@navercorp.com }
\and
Dongyoon Han\footnotemark[1]\\
{\tt\small dongyoon.han@navercorp.com}
\and
Sangdoo Yun\footnotemark[1]\\
{\tt\small sangdoo.yun@navercorp.com }
}

\maketitle

\begin{abstract}
In this paper, we propose a new multi-scale face detector having an extremely tiny number of parameters (EXTD), less than $0.1$ million, as well as achieving comparable performance to deep heavy detectors. 
While existing multi-scale face detectors extract feature maps with different scales from a single backbone network, our method generates the feature maps by iteratively reusing a shared lightweight and shallow backbone network. 
This iterative sharing of the backbone network significantly reduces the number of parameters, and also provides the abstract image semantics captured from the higher stage  of the network layers to the lower-level feature map.
The proposed idea is employed by various model architectures and evaluated by extensive experiments.
From the experiments from WIDER FACE dataset, we show that the proposed face detector can handle faces with various scale and conditions, and achieved comparable performance to the more massive face detectors that few hundreds and tens times heavier in model size and floating point operations.

\end{abstract}
\section{Introduction}
\label{sec:introduction}
Detecting faces in an image is considered to be one of the most practical tasks in computer vision applications, and many studies~\cite{viola2001rapid,mathias2014face} are proposed from the beginning of the computer vision research. 
After the advent of deep neural networks, many face detection algorithms~\cite{yang2016wider,zhang2017s3fd,tian2018learning, wang2017detecting} applying the deep network have reported significant performance improvement to the conventional face detectors. 

The state-of-the-art (SOTA) face detectors~\cite{zhang2017s3fd,tian2018learning, wang2017detecting} for in-the-wild images employ the framework of the recent object detectors~\cite{girshick2015fast,ren2015faster,redmon2017yolo9000,redmon2018yolov3,liu2016ssd,dai2016r,lin2017feature}. 
These methods can even handle a various scale of faces with difficult conditions such as distortion, rotation, and occlusion. Among them, the face  detectors~\cite{zhang2017s3fd,najibi2017ssh,yang2017face,tang2018pyramidbox,chi2018selective,zhang2018single} using multiple feature-maps from different layer locations, which mainly stem from \cite{liu2016ssd,lin2017feature,lin2017focal}, are dominantly used since these methods can handle the faces with various scale in a single forward path.

\begin{figure}[t]
\begin{center}
\includegraphics[width=0.99\linewidth]{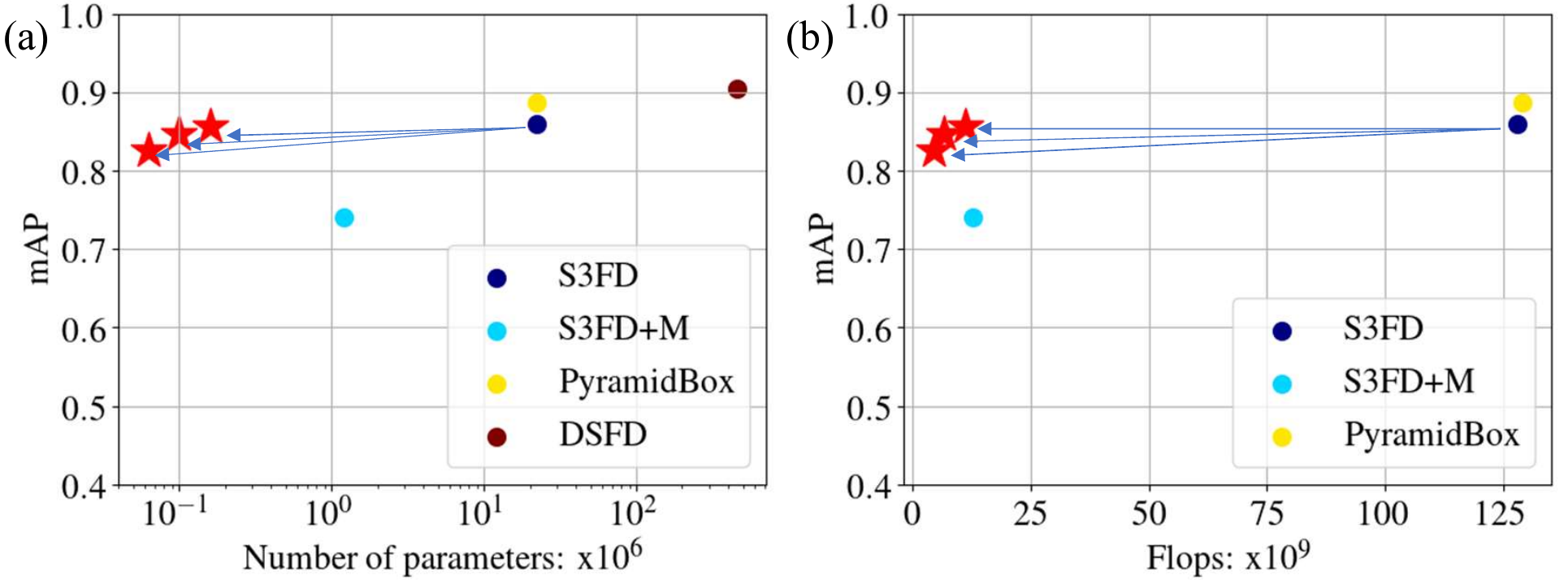}
\end{center}
   \caption{Illustration of the mean average precision (mAP) regarding the parameter size (a) and Flops (b) evaluated on WIDER FACE dataset. Our method (star) shows comparable mAP to S3FD~\cite{zhang2017s3fd} with a significantly smaller model.
   Red stars denote the proposed models with various sizes.
   `S3FD+M' denotes the S3FD variation using MobileFaceNet~\cite{chen2018mobilefacenets} as a backbone network instead of VGG-16~\cite{simonyan2014very}.
   Best viewed in wide and colored vision.}
\label{fig:teaser}
\vspace{-2mm}
\end{figure}

While these methods achieved impressive detection performance, they commonly share two problems. 
One is their large number of parameters.
Since they use a large classification network such as VGG-16~\cite{simonyan2014very}, ResNet~\cite{he2016deep}-50 or 101, and DenseNet-169~\cite{huang2017densely}, the number of total parameters exceed $20$ million, over $80$Mb supposing $32$-bit floating point for each parameter.
Furthermroe, the amount of floating point operations (FLOPs) also exceeds $100$G, and these make it nearly impossible to use the face detectors in CPU or mobile environment, where the most face applications run in.
The second problem, from the architecture perspective, is the limited capacity of the low-level feature map in capturing object semantics.
The most single-shot detector (SSD)~\cite{liu2016ssd} variant object and face detectors struggle the problem because the low-level feature map passes shallow convolutional layers.
To alleviate the problem, the variants of Feature pyramid network (FPN) architecture such as \cite{liu2016ssd,lin2017feature,shen2017dsod,sun2018fishnet} are used but requires additional parameters and memories for re-expanding the feature map.

In this paper, we propose a new multi-scale face detector with extremely tiny size (EXTD) resolving the two mentioned problems. 
The main discovery is that we can share the network in generating each feature-map, as shown in Figure~\ref{fig:teaser}. 
As in the figure, we design a backbone network such that reduces the size of the feature map by half, and we can get the other feature maps with recurrently passing the network. 
The sharing can significantly reduce the number of parameters, and this enables our model to use more layers to generate the low-level feature maps used for detecting small faces.
Also, the proposed iterative architecture makes the network to observe the features from various scale of faces and from various layer locations, and hence offer abundant semantic information to the network, without adding additional parameters. 

Our baseline framework follows FPN-like structures, but can also be applied to SSD-like architecture.
For SSD based architecture, we adopt the setting from \cite{zhang2017s3fd}.
For the FPN architectures, we refer an up-sampling strategy from \cite{li2018tiny}.
The backbone network is designed to have less than $0.1$ million parameters with employing inverted residual blocks proposed in MobileNet-V2~\cite{sandler2018mobilenetv2}.
We note that our model does not require any extra layer commonly defined as in \cite{liu2016ssd,lin2016feature}, and is trained from scratch.
We evaluated the proposed detector and its variants on WIDER FACE~\cite{yang2016wider} dataset, the most widely used and similar to the in-the-wild situation.

The main contributions of this work can be summarized as follows: 
(1) We propose an iterative network sharing model for multi-stage face detection which can significantly reduce the parameter size, as well as provide abundant object semantic information to the lower stage feature maps. 
(2) We design a lightweight backbone network for the proposed iterative feature map generation with $0.1$M number of parameters, which less than 400Kb, and achieved comparable mAP to the heavy face detection methods.
(3) We employ the iterative network sharing idea to the widely used detection architectures, FPN and SSD, and show the effectiveness of the proposed scheme.

\section{Related Works}
\label{sec:related_works}

\noindent\textbf{Face detectors: }
Face detection has been an important research topic since an initial stage of computer vision researches.  
Viola~\etal~\cite{viola2001rapid} proposed a face detection method using Haar features and Adaboost with decent performance, and several different approaches~\cite{li2002statistical,mita2005joint,yang2014aggregate,mathias2014face} followed.
After deep learning has become dominant, many face detection methods applying the techniques have been published.
In the early stages, various attempts were tried to employ the deep architecture to face detection, such as cascade architecture~\cite{yang2016wider,zhang2016joint}, and occlusion handling~\cite{yang2015facial}.

Recent face detectors has been designed based on the architecture of generic object detectors including Faster-RCNN~\cite{ren2015faster}, R-FCN~\cite{dai2016r}, SSD~\cite{liu2016ssd}, FPN~\cite{lin2017feature}, and RetinaNet~\cite{lin2017focal}.
Face RCNN and its variants~\cite{wang2017faceb,jiang2017face,zhang2018face} apply Faster-RCNN, and \cite{wang2017detecting,zhu2016cms} use R-FCN for detecting faces with meaningful performance improvements.

Also, to cope with the various scale of faces with single forward path, object detectors such as SSD, RetinaNet, and FPN are dominantly adopted since they use features from multiple layer locations for detecting objects with various scale in a single forward path.
S3FD~\cite{zhang2017s3fd} achieved promising performance by applying SSD with introducing multiple strategies to handle the small size of faces.
FAN~\cite{wang2017face} uses RetinaNet by applying anchor level attention to detect the occluded faces.
After S3FD, many improved versions~\cite{tang2018pyramidbox, yang2017face,zhu2018seeing,li2018dsfd,zhang2018single} are introduced and achieved performance gain from the previous methods.
FPN based face detection methods~\cite{chi2018selective,zhang2019improved,tian2018learning} achieved SOTA performance by enhancing the expression capacity of the lower-level feature map used for detecting small faces.

The mentioned SOTA methods commonly use classification network such as VGG-16~\cite{simonyan2014very}, ResNet-50 or 101~\cite{he2016deep}, and DenseNet-169~\cite{huang2017densely} as a backbone of the model.
These classification networks have a large number of parameters exceeding $20$ million, and the model size is over $80$Mb supposing $32$-bit floating point for each parameter.
Some cascade methods such as \cite{yu2019anchor} report decent mAP with the smaller mount of model size, about $3.8$Mb.
However, the size is still burdensome to the devices like mobile, because users generally want their applications not to exceed few ten's of Mb.
Also, the face detector should mostly be much smaller than the total size of the application because a face detector is usually an end-level function of the application. 

Here, we propose a new scheme of iteratively sharing the backbone network, which can be applicable to both SSD and FPN based architectures. The method achieves comparable accuracy to the original models, and the overall model size is extremely smaller as well.

\noindent\textbf{Lightweight generic object detectors: }
Recently, for detecting general objects in condition with a limited resource such as mobile devices, various single-stage, and two-stage lightweight detectors were proposed. For the single-stage detectors, MobileNet-SSD~\cite{howard2017mobilenets}, MobileNetV2-SSDLite ~\cite{sandler2018mobilenetv2}, Pelee~\cite{wang2018pelee} and Tiny-DSOD~\cite{li2018tiny} were proposed. For two-stage detectors, Light-Head R-CNN~\cite{li2017light} and ThunderNet~\cite{qin2019thundernet} were proposed.
The mentioned methods achieved meaningful accuracy and size trade-off, but we aim to develop a detector which has a much smaller number of parameters with introducing a new paradigm, iterative use of the backbone network.

\noindent\textbf{Recurrent convolutional network: }
The idea of recurrently using convolutional layers has been applied to various computer vision applications.  Sharesnet~\cite{boulch2017sharesnet} and Iamnn~\cite{leroux2018iamnn} applied recurrent residual network into classification task. 
Guo~\etal~\cite{guo2019depthwise} reduce the parameters by sharing depthwise convolutional filters in learning multiple visual domain data.
The iterative sharing is also applied to dynamic routing~\cite{kemaev2018reset}, fast inference of video~\cite{pan2018recurrent}, feature transfer~\cite{liu2017efficient}, super-resolution~\cite{kim2016deeply}, and recently in segmentation~\cite{li2019dfanet}. 
In this paper, we introduce a method applying the concept of iterative convolutional layer sharing in the face detection task, which is the first to the best of our knowledge.

\begin{figure*}[t]
\begin{center}
\includegraphics[width=0.95\linewidth]{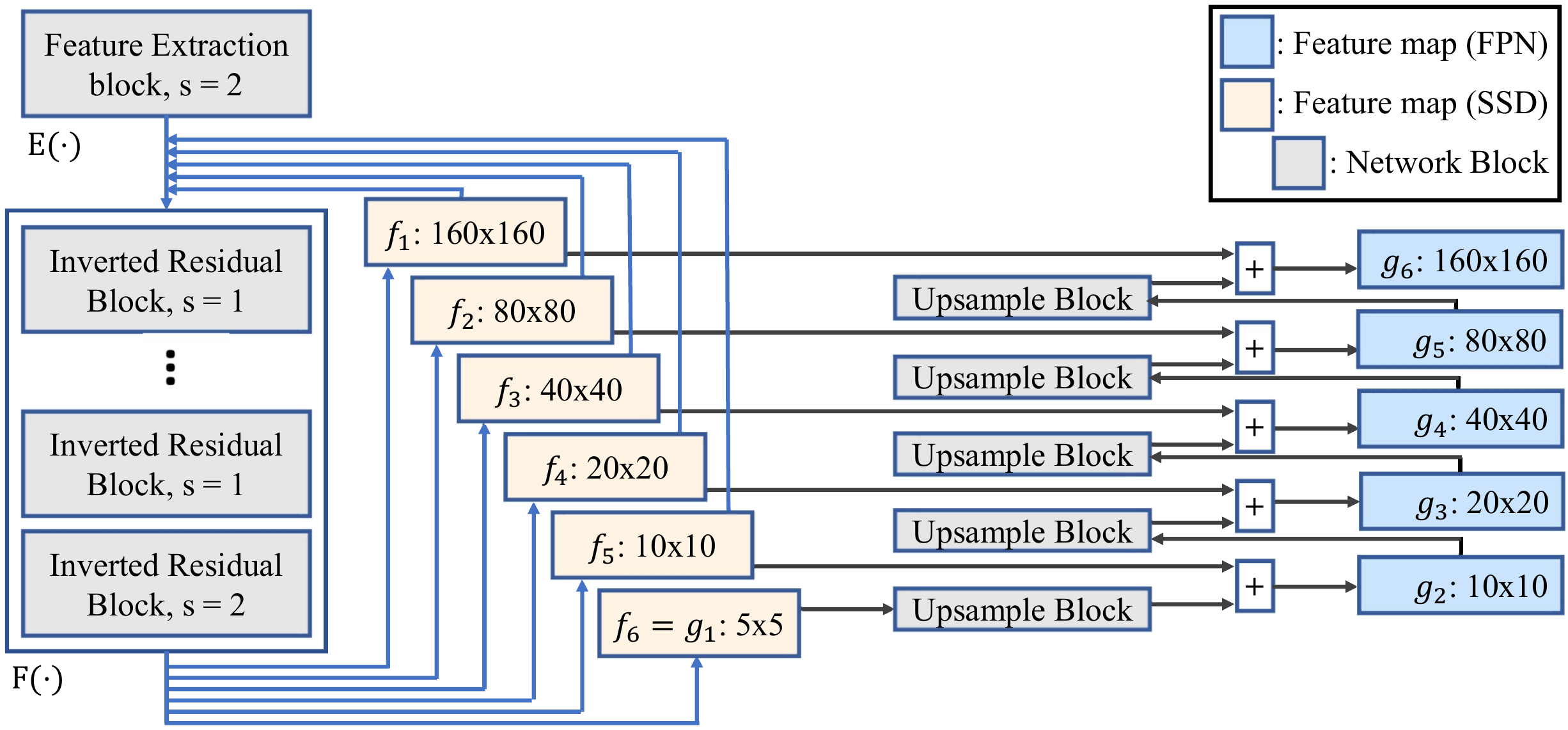}
\end{center}
\vspace{-2mm}
   \caption{The overall framework of the proposed method. The structure recurrently generates the feature maps $f_i$ (SSD version), and we upsample the feature maps with skip connection to generate the feature maps $g_i$ (FPN version). The classification and regression heads can be attached to either $f_i$ and $g_i$. }
\label{fig:teaser}
\end{figure*}

\section{EXTD}
\label{sec:proposed_method}
In this section, we introduce the main components of the proposed work including iterative feature map generation, the architectures of the proposed face detection models, backbone networks, and classification and regression head design.
Also, implementation details for designing and training the models will be introduced.

\subsection{Iterative Feature Map Generation}
\label{sec:RFMap}
Figure~\ref{fig:teaser} shows the overall framework of the  proposed method with two variations, SSD-like, and FPN-like frameworks.
In the proposed method, we get multiple feature maps with different resolutions by recurrently passing the backbone network. Let assume that $F(\cdot)$ and $E(\cdot)$ each denotes the backbone network and the first Conv layer with stride two. Then, the iterative process is defined as follows:
\begin{eqnarray}
\begin{aligned}
\label{eq:recurrent_simple}
f_i = F(f_{i-1}&),~i = 1,...,N,\\
f_0& = E(x). \\
\end{aligned}
\end{eqnarray}
Here, the set $\{f_1,..,f_N\}$ denote the set of feature maps, and $x$ is the image.
In FPN version, we upsample each feature map and connect the previous feature maps via skip-connection~\cite{he2016deep, ronneberger2015u}. 
The upsampling step $U_i(\cdot)$ is conducted with bilinear upsampling followed by an upsampling block composed of separable convolution and point-wise convolution, inspired by \cite{li2018tiny}.
The resultant set of the feature map $G = \{g_1,...,g_N\}$ is obtained as,

\begin{eqnarray}
\begin{aligned}
\label{eq:upsampling}
g_{i+1} = U_i(g_{i}) + &f_{N-i},~i = 1,...,N-1,\\
g_1& = f_N. \\
\end{aligned}
\end{eqnarray}

For the SSD-like architecture, which is the first variant, we extract feature maps $f_i$ and connect the classification and regression head to the feature maps. 
In FPN-like architecture, the feature maps $g_i$ from equation (\ref{eq:upsampling}) are used.
The classification and regression heads are designed by a 3x3 convolutional network and hence, both models are designed as a fully convolutional network. 
This enables the models to deal with various size of images. The detailed implementation of the heads is introduced in below sections.

For all the cases, we set the image $x$ to have 640x640 resolution in training phase and use $N=6$ number of feature maps. 
Hence, we get 160x160, 80x80, 40x40, 20x20, 10x10 and 5x5 resolution feature maps. 
In each location of the feature map, prior anchor candidates for the face is defined, following the same setting as S3FD~\cite{zhang2017s3fd}.

\begin{figure}[t]
\begin{center}
\includegraphics[width=0.95\linewidth]{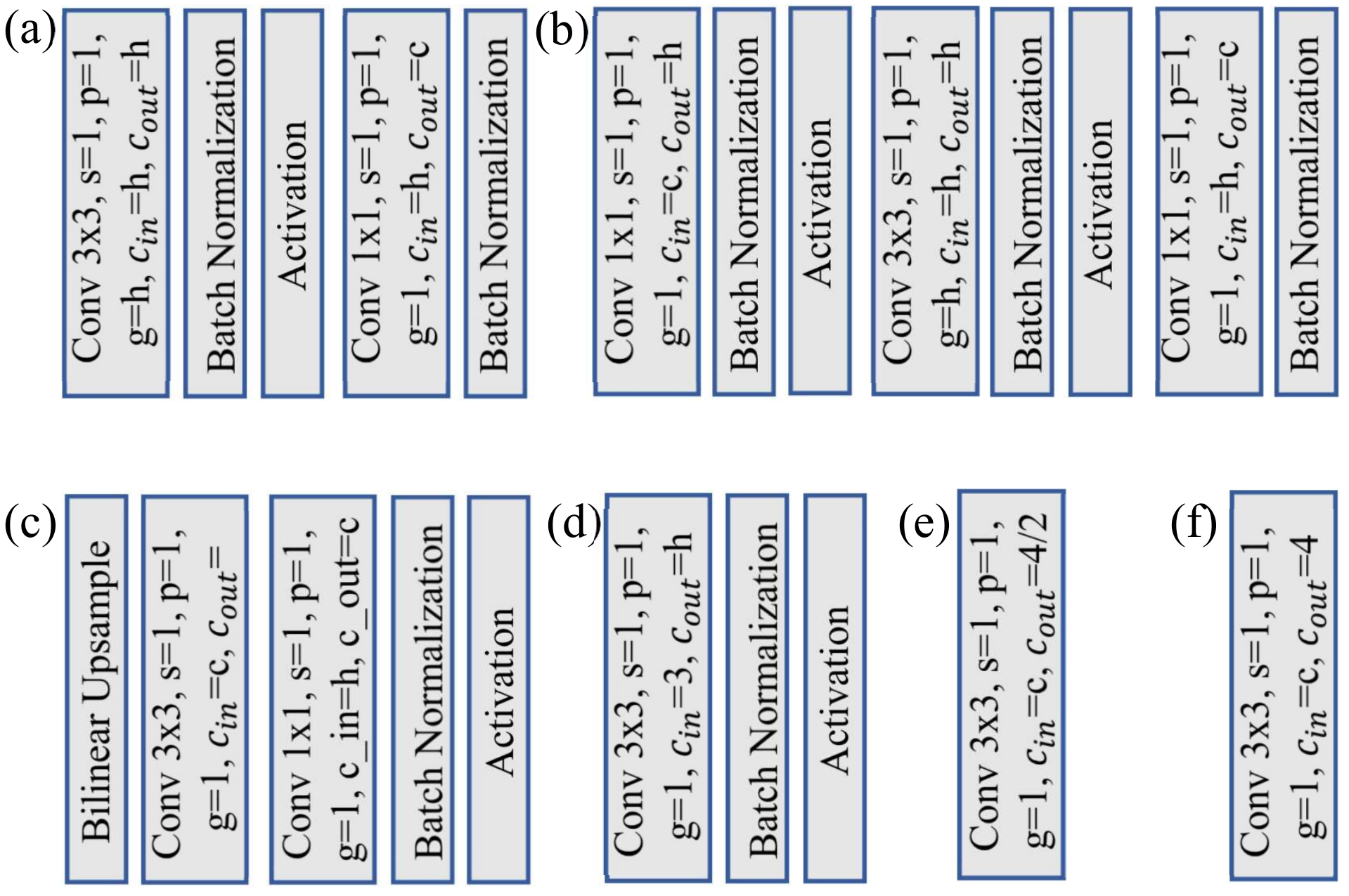}
\end{center}
   \caption{Detailed configuration of the components.
   The terms s, p, g, $c_{in}$, and $c_{out}$ denote the stride, padding, group, input channel width, and output channel width.
   Figures (a) and (b) each shows the initial and remaining inverted residual blocks. In (c) and (d), upsampling block and the Feature extraction block are presented. Figures (e) and (f) each denotes the classification and regression head. For the activation function, PReLU or Leaky-ReLU are used for (a) and (b), and ReLU is used for the others.}
\label{fig:backbone}
\end{figure}

One notable property of this architecture is that this method provides more abundant semantic information in lower-level feature maps compared to the face detectors adopting SSD architecture. 
While the existing methods commonly report the problem that the lower-level feature maps only contain limited semantic information due to their limited length of depth, our iterative architecture repeatedly shows intermediate level features and the various scale of faces to the network.
We conjecture that the different features have similar semantics because the target objects in our case are faces, and the faces share homogeneous shapes regardless of their scale dissimilar to general objects.
In Section~\ref{sec:experiments}, we show that the proposed method clearly enhances the detection accuracy for small size faces, and this can be more improved  by taking the FPN architecture.

\subsection{Model Component Description}
In the proposed model, a lightweight backbone network reducing the feature map resolution by half is used. 
The network is composed of inverted residual blocks followed by one 3x3 convolutional (Conv) filter with stride 2, based on \cite{sandler2018mobilenetv2,chen2018mobilefacenets}. 
The inverted residual block is composed of a set of point-wise Conv, separable Conv, and point-wise Conv. 
In each block, the channel width is expanded in the first point-wsie Conv and then, squeezed by the last point-wise Conv filter. 
The default setting of the network depth is set to $6$ or $8$, and the output channel width is set to $32$,$48$ or $64$, which do not largely exceed overall $0.1$ million parameters. 
Different from MobileNet-V2~\cite{sandler2018mobilenetv2}, PReLU~\cite{he2015delving} (or leaky-ReLU) is applied and shown to be more successful than ReLU in training the proposed recurrent architecture.
This phenomenon will be further discussed in Section~\ref{sec:experiments}.

Other than the inverted residual block, the proposed architecture also includes feature extraction block, upsampling blocks, and classification and regression heads.
The detailed description of the components is introduced in Figure~\ref{fig:backbone}.
The figures in (a) and (b) each shows the inverted residual block architecture.
Residual skip-connection is applied when the input and output channel width are equivalent, and at the same time, the stride is set to one.
The upsampling block in (c) consists of bilinear upsample layer followed by depth-wise and point-wise Conv blocks.
Feature extraction block (d) is defined by a 3x3 Conv network followed by batch normalization and the activation function. 
The classification (e) and regression (f) heads are also defined by a 3x3 Conv network.
The implementation of the head is described in Section~\ref{sec:head}.

\subsection{Classification and Regression Head Design}
\label{sec:head}
For detecting the faces using the generated feature maps, we use a classification head and a regression head for each feature map to classify whether each prior box contains a face, and to regress the prior box to the exact location. 
The classification and regression heads are both defined as single $3$x$3$ Conv filters as shown in Figure~\ref{fig:backbone}.
The classification head $C_i$ has two-dimensional output channel $c_i$ except $C_1$ that having four-dimensional channels.
For $C_1$, we apply Maxout~\cite{goodfellow2013maxout} approach to select two of the four channels for alleviating the false positive rate of the small faces, as introduced in S3FD.
The regression head $R_i$ is defined to have output feature $r_i$ to have four-dimensionional channel, and each denotes width, height ratio, and center locations, adopting the dominantly used setting in RPN~\cite{ren2015faster}.

\subsection{Training}
The proposed backbone network and the classification and regression head are jointly trained by a multitask loss function from RPN composed of a classification loss $l_c$ and a regression loss $l_r$ as, 
\begin{eqnarray}
\begin{aligned}
\label{eq:loss}
l(\{c_j,r_j\}) = \frac{\lambda}{N_{cls}}\sum_j l_{c}(c_j, {c^*_j}) + \frac{1}{N_{reg}}\sum_j{c^*_j}l_{r}(r_j, {r^*_j})
\end{aligned}
\end{eqnarray}
Here, $j$ is the index of the anchor boxes, and the label $c^*_j\in\{0,1\}$ and $r^*_j$ is the ground truth of the anchor box. The label $c^*_j$ is set to $1$ when Jaccard overlap~\cite{erhan2014scalable} between the anchor box and ground trurh box is higher than a threshold $\textit{t}$. 
The denominator $N_{cls}$ denotes the total number of positive and negative samples.
The regression loss is computed only for the positive sample and hence, the number $N_{reg}$ is defined by $N_{reg} = \sum_j{c^*_j}$.
The parameter $\lambda$ is defined to balance the two losses because $N_{cls}$ and $N_{reg}$ are different from each other.
The vector $r^*_j$ denotes the ground truth box location and size for the face.
The classification loss $l_c$ and the regression loss $l_r$ are defined as cross-entropy loss and smooth-$\ell1$ loss, respectively.

The primary obstacle for the classification in the face detection task is a class imbalance problem between the face and the background, especially regarding the small faces. To alleviate the problem, we also adopt the strategies including online hard negative mining and scale compensation anchor matching introduced in S3FD. Using the hard negative mining technique, we balance the ratio of positive and negative samples ${N_{neg}}/{N_{pos}}$ to $3$ and the balancing parameter $\lambda$ is set to $4$.
Also, from the scale compensation anchor matching strategy, we first pick the positive samples where the Jaccard overlap is over $0.35$, and then further pick the remaining samples in sorted order from  the samples that their Jaccard overlap is larger than $0.1$ if the number of positive samples is insufficient.

For Data augmentation, we follow the conventional augmentation setting from S3FD.
The augmentation includes color distortions~\cite{howard2013some}, random crop, horizontal flip, and vertical flip.
The proposed method is implemented with PyTorch~\cite{paszke2017automatic} and NAVER Smart Machine Learning (NSML)~\cite{kim2018nsml} system.
Please refer Appendix~\ref{app:implementation} to see the detailed training and optimization settings for training the proposed network.
Code will be available at \url{https://github.com/clovaai}.

\begin{table*}[]
\small
\centering
\tabcolsep=0.3cm
\begin{tabular}{@{}lcccccc@{}}
\toprule
\multirow{2}{*}{Model} & 
\multirow{2}{*}{Backbone} &
\multirow{2}{*}{\# Params} & 
\multirow{2}{*}{\# Madds (G)} &
\multirow{2}{*}{\begin{tabular}[c]{@{}c@{}} \\ Easy (mAP)\end{tabular}} &
\multirow{2}{*}{\begin{tabular}[c]{@{}c@{}}WIDER FACE \\ Medium (mAP) \end{tabular}} &
\multirow{2}{*}{\begin{tabular}[c]{@{}c@{}} \\ Hard (mAP) \end{tabular}} \\
                                  &
                                  &                            
                                  &
                                  &                                  
                                  &        \\ \midrule
PyramidBox~\cite{tang2018pyramidbox}*                      & VGG-16 & 57 M & 129 & 0.961 / 0.956   & 0.950 / 0.946& 0.887 / 0.887\\
DSFD~\cite{li2018dsfd}-ResNet101*  & ResNet101  & 399 M & - & 0.963  & 0.954  & 0.901 \\
DSFD-ResNet152*  & ResNet152  & 459 M & - & 0.966 / 0.960  & 0.957 / 0.953  & 0.904 / 0.900\\
S3FD~\cite{zhang2017s3fd}*                      & VGG-16 & 22 M & 128 & \textbf{0.942 / 0.937}   & \textbf{0.930 / 0.925}& \textbf{0.859 / 0.858}\\
S3FD - Scratch                      & VGG-16 & 22 M & 128 & 0.931    & 0.920 & 0.846 \\
S3FD + MobileFaceNet~\cite{chen2018mobilefacenets}  & MobileFaceNet  & 1.2 M & 12.7 & 0.881  & 0.859  & 0.741 \\
\midrule
EXTD-FPN-32-PReLU      & -  & 0.063 M & 4.52 & 0.896   & 0.885    & 0.825\\ 
EXTD-FPN-48-PReLU      & -  & 0.10 M   & 6.67  & 0.913    & 0.904    & 0.847\\ 
\textbf{EXTD-FPN-64-PReLU}      & -  & 0.16 M   & 11.2  & \textbf{0.921 / 0.912}    & \textbf{0.911 / 0.903}    & \textbf{0.856 / 0.850}\\ 
\midrule
\end{tabular}
\caption{Quantitative comparison to recent state-of-the-art face detectors on WIDER FACE dataset. `*' denotes results reported in the original papers. 
For the proposed model with highest validation mAP, we list the mAPs from validation set and that from test set at the left-side and right-side of the slash in fifth to seventh columns. The other cases, mAPs from the validation set are listed.
}
\vspace{-2mm}
\label{table:wider_face_comp}
\end{table*}
\begin{figure*}[t]
\begin{center}
\subfigure[Validation Easy]{\includegraphics[width=0.32\linewidth]{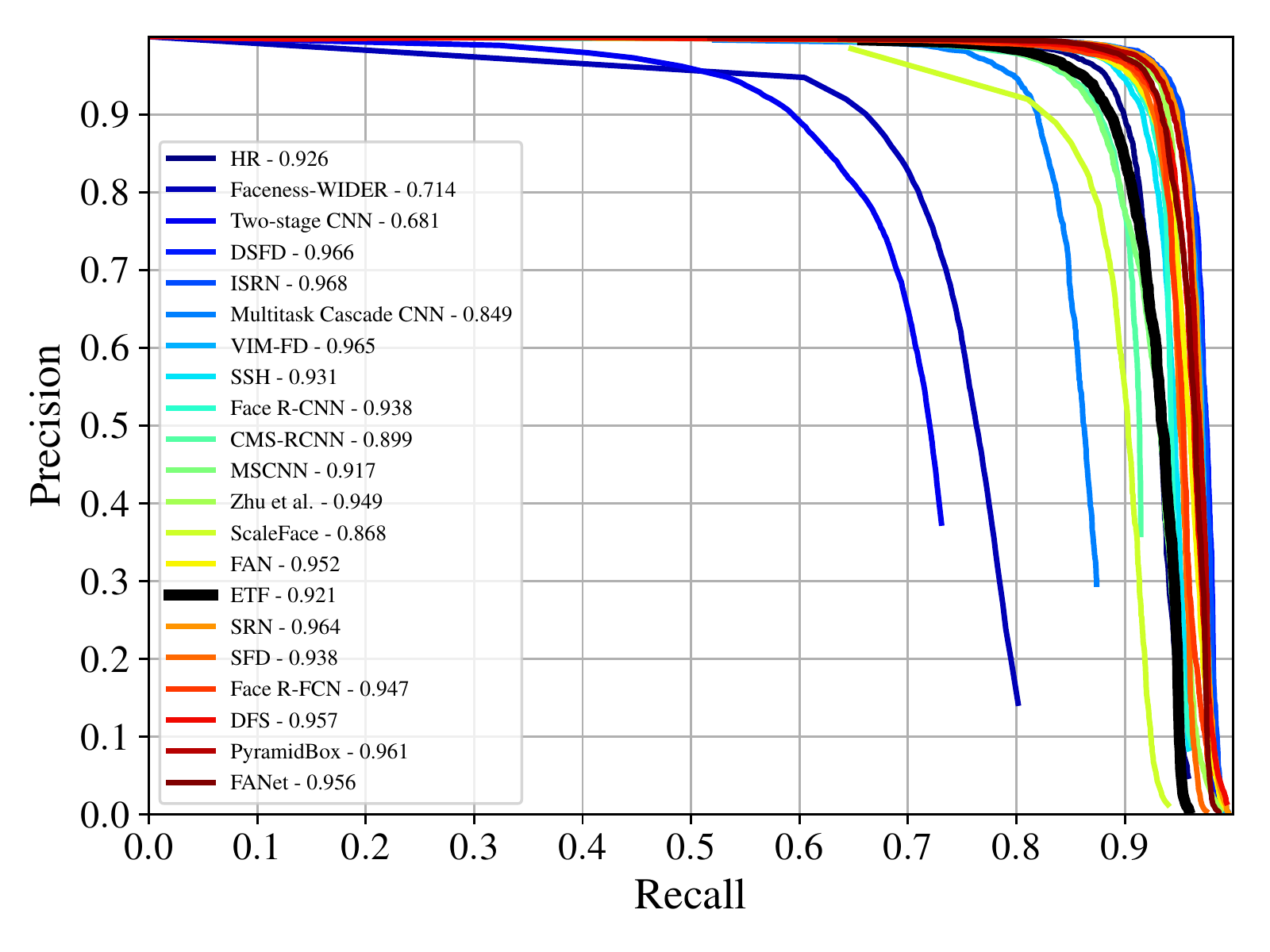}}
\subfigure[Validation Medium]{\includegraphics[width=0.32\linewidth]{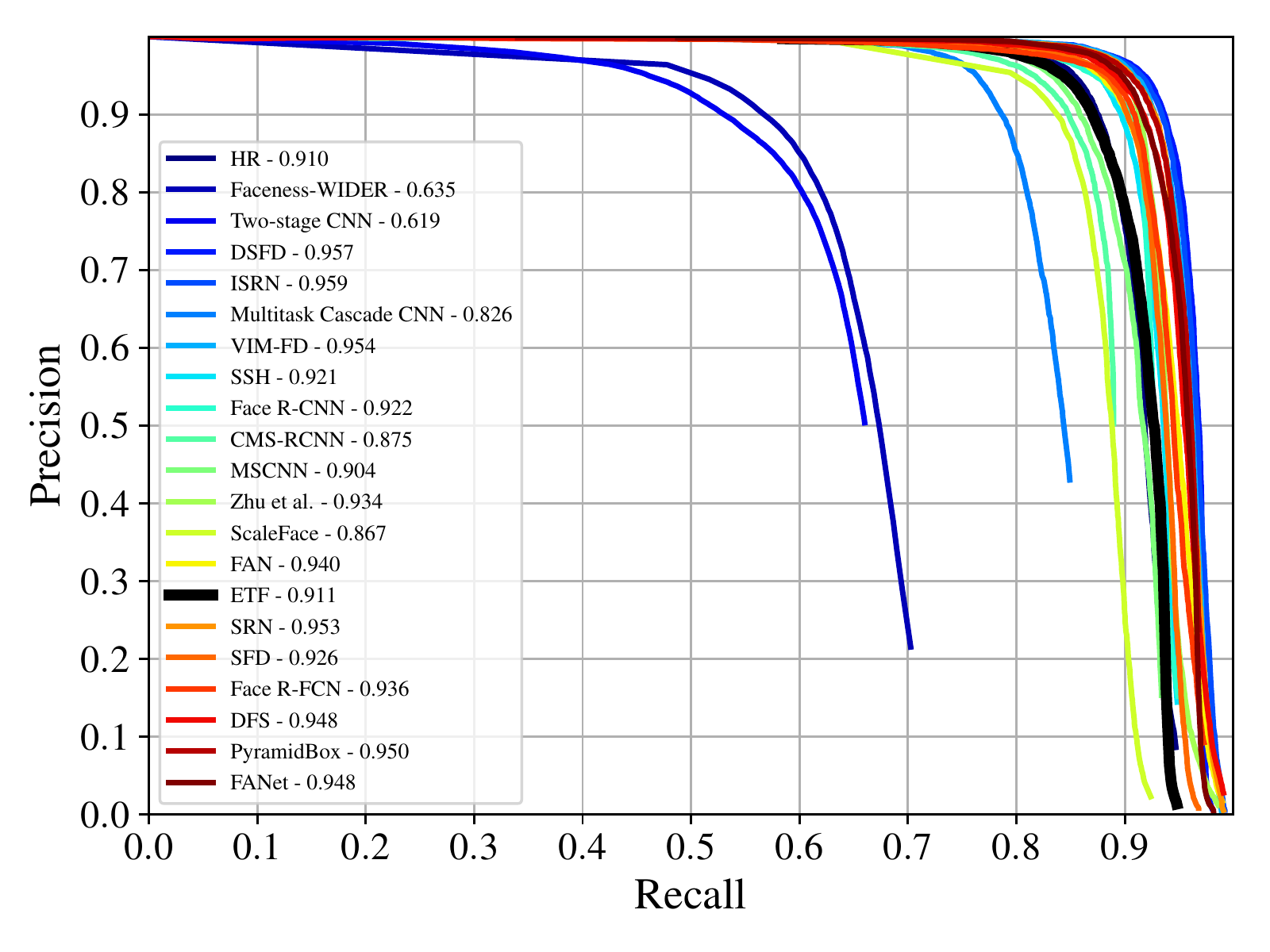}}
\subfigure[Validation Hard]{\includegraphics[width=0.32\linewidth]{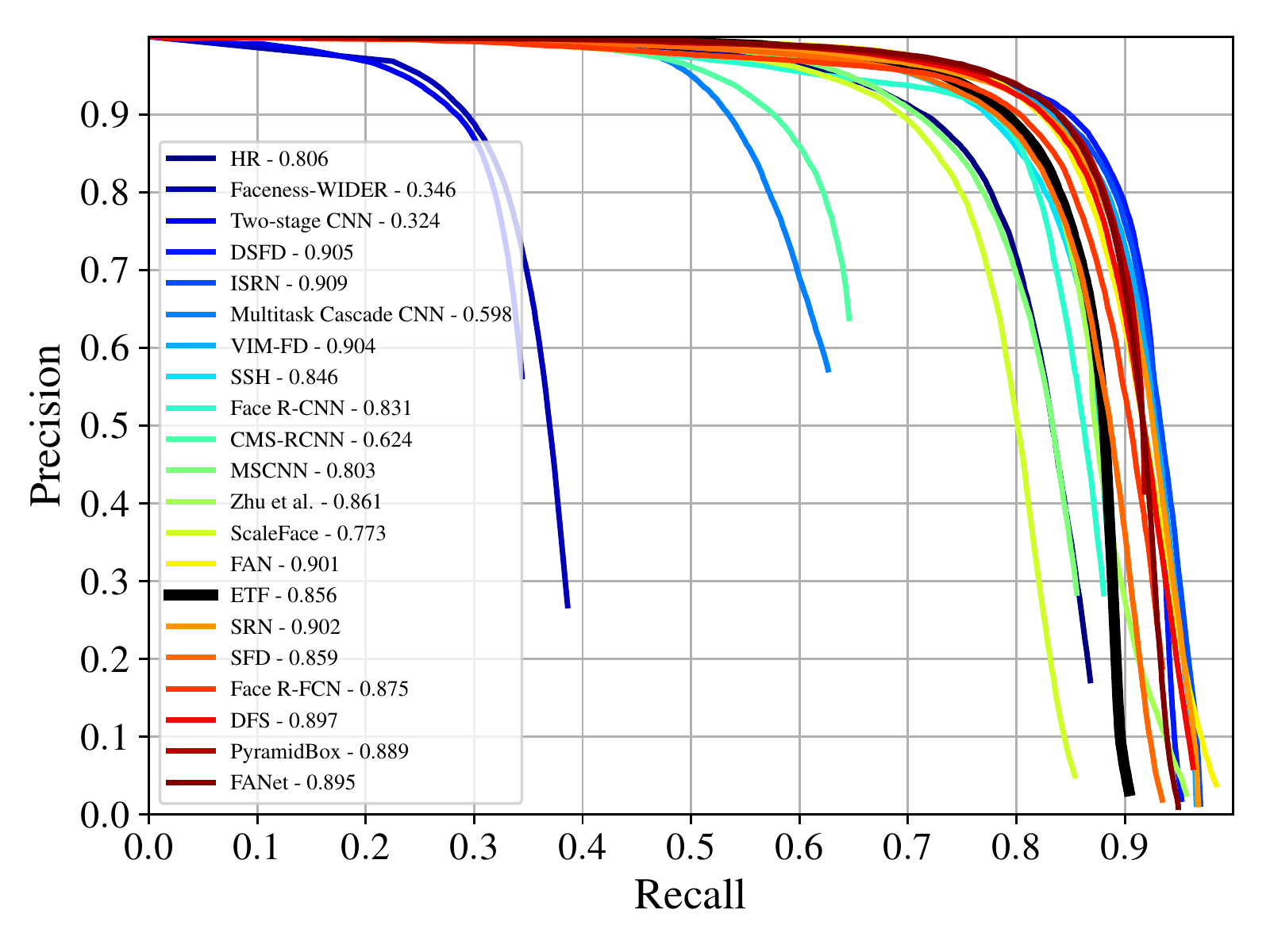}}
\subfigure[Test Easy]{\includegraphics[width=0.32\linewidth]{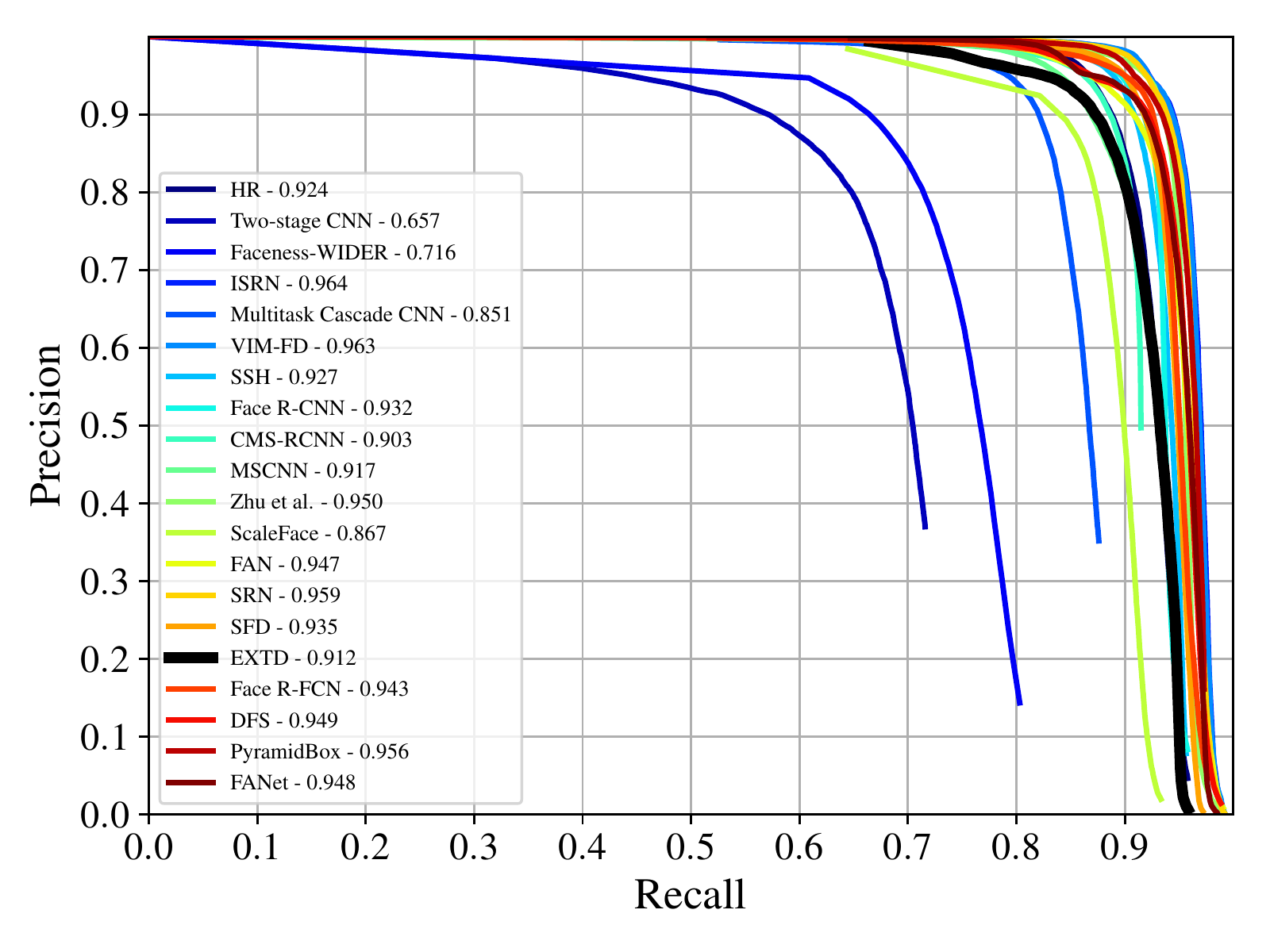}}
\subfigure[Test Medium]{\includegraphics[width=0.32\linewidth]{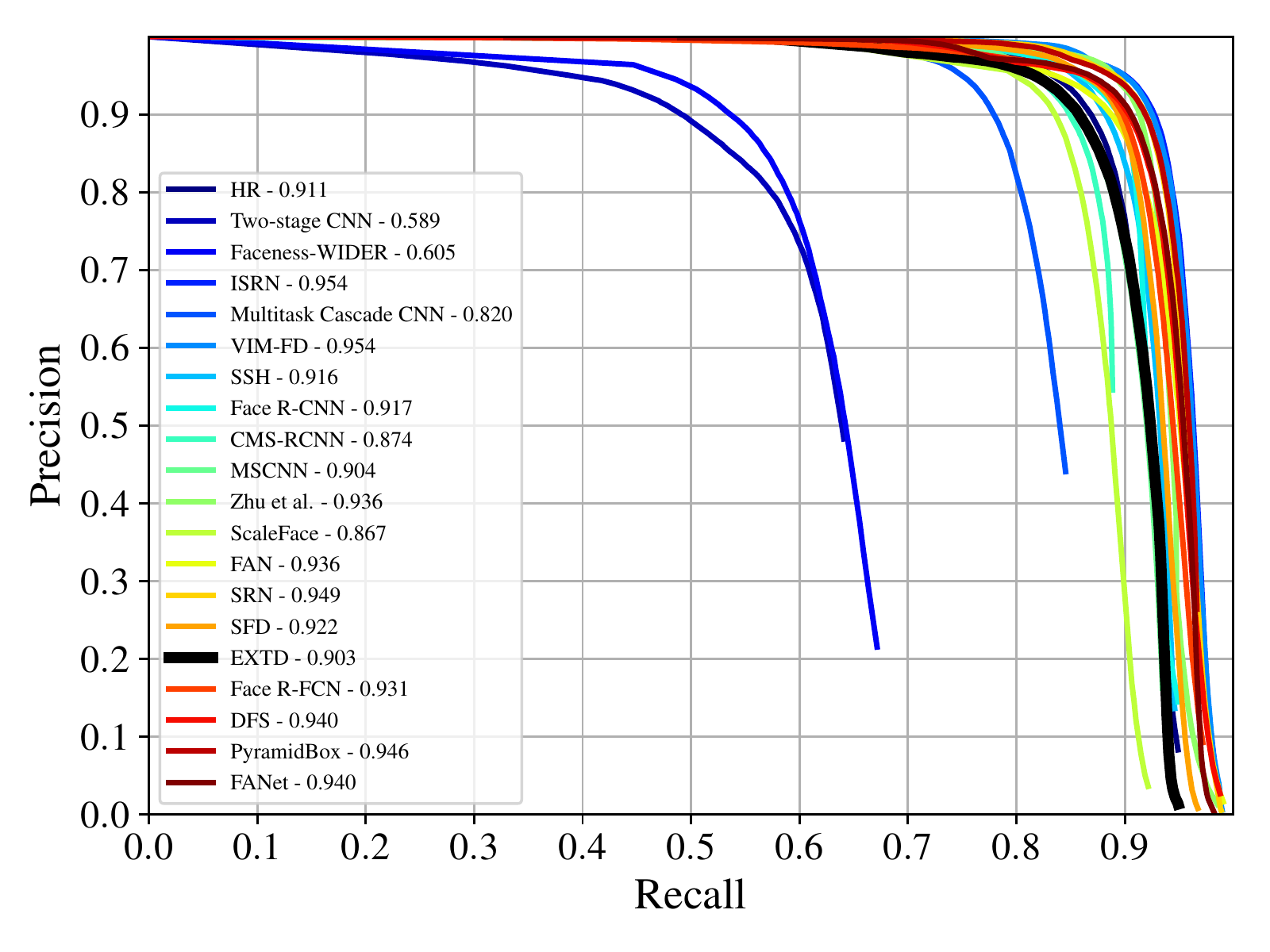}}
\subfigure[Test Hard]{\includegraphics[width=0.32\linewidth]{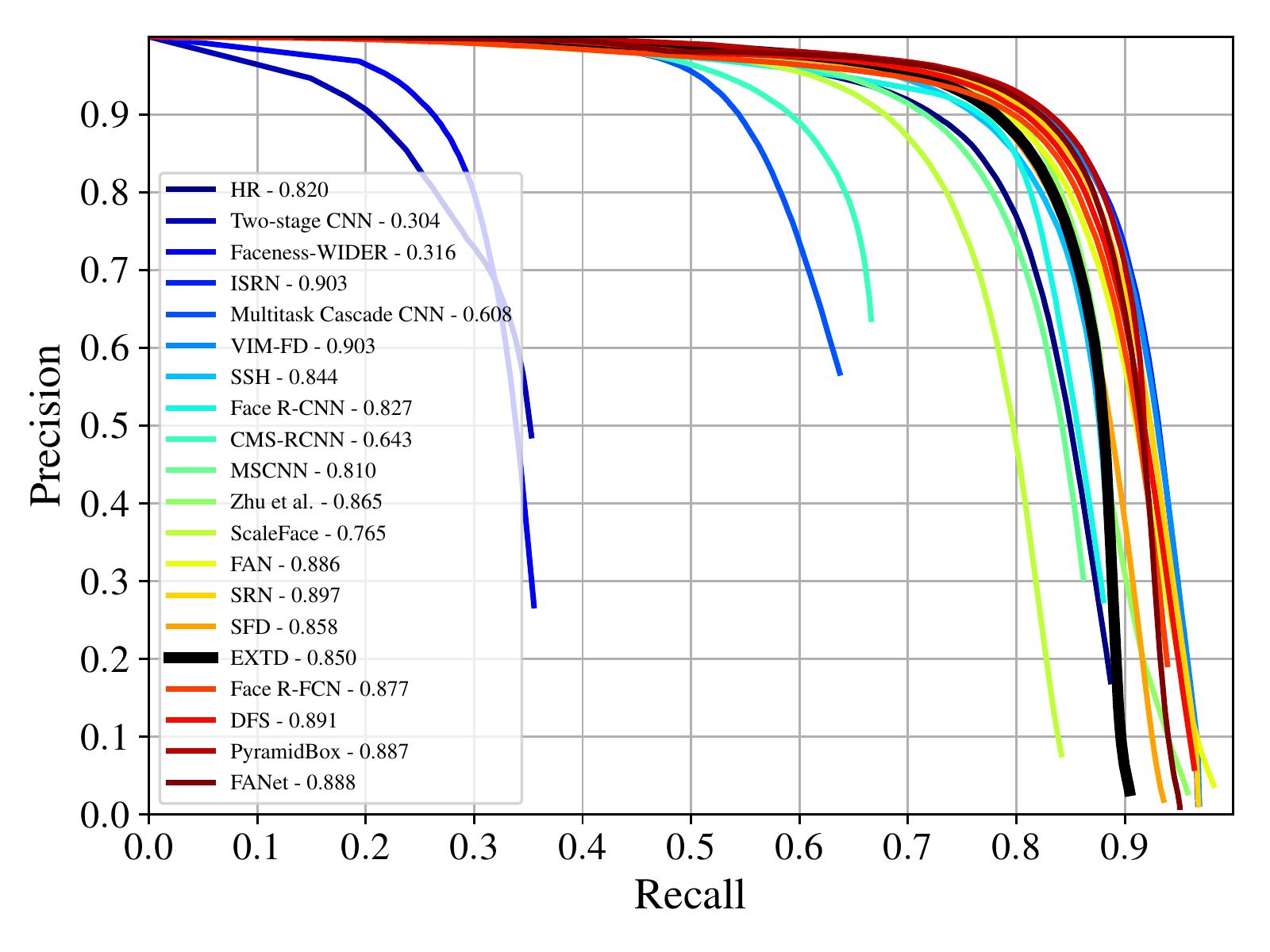}}
\end{center}
\vspace{-3mm}
   \caption{ROC curves on WIDER FACE dataset. Best viewed in wide vision. The curves from our method are illustrated by `black'.}
\label{fig:roc_curve}
\vspace{-3mm}
\end{figure*}

\section{Experiments}
\label{sec:experiments}
In this section, we quantitatively and qualitatively analyze the proposed method with various ablations.
For the quantitative analysis, we compare the detection performance of the proposed method and the SOTA face detection algorithms. 
Qualitatively, we show that our method can successfully detect faces in various conditions. 

\begin{figure*}[t]
\begin{center}
\includegraphics[width=0.99\linewidth]{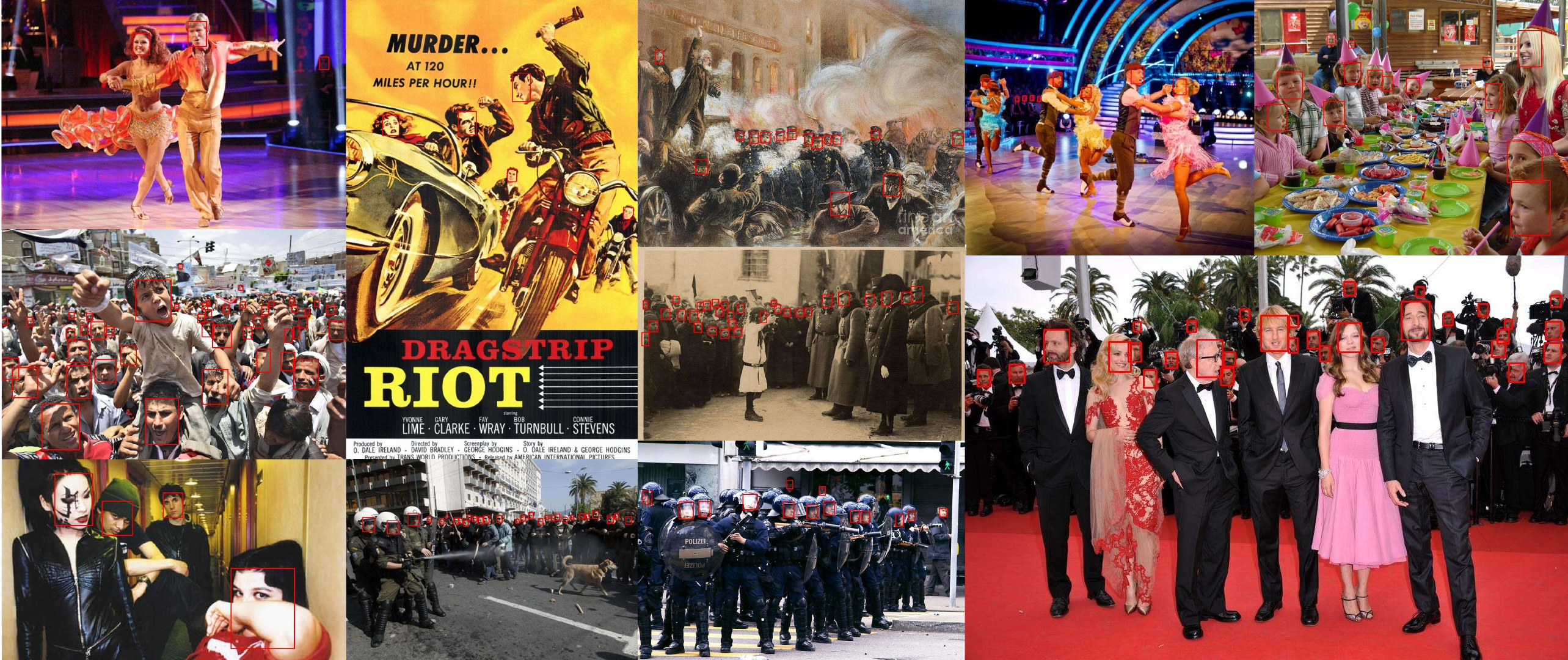}
\end{center}
\vspace{-2mm}
   \caption{Illustration of the face detection results. The illustration includes vulnerable cases such as scale, illumination, face print, occlusion, pose, color, and paintings. \textit{EXTD-FPN-64-PReLU} version was used to detect the images. Best viewed in wide and colored vision.}
\label{fig:qual}
\vspace{-2mm}
\end{figure*}

\subsection{Experimental Setting}
\textbf{Datasets:} 
we tested the proposed method and ablations of the method with WIDER FACE~\cite{yang2016wider} dataset, which is most recent and is similar to in-the-wild face detection situation.
The images in the dataset are divided into Easy, Medium, and Hard cases which are roughly categorized by different scales: large, medium, and small, of faces. 
The Hard case includes all the images of the dataset, and the Easy and Medium cases both are the subsets of the Hard case.
The dataset has total 32,203 images with 393,703 labeled faces and is split into training (40$\%$), validation (20$\%$) and testing (40$\%$) set.
We trained the detectors with the training set and evaluated them with validation and test  sets.

\textbf{Comparison: } 
Since our method followed the training and implementation details such as anchor design, data augmentation, and feature-map resolution design equivalent to S3FD~\cite{zhang2017s3fd}, which has become one of the baseline methods in face detection field, we mostly evaluated the performance by comparing the S3FD model and its SOTA variations~\cite{tang2018pyramidbox,li2018dsfd}.
The other techniques based on the S3FD model such as Pyramid anchor~\cite{tang2018pyramidbox}, Feature enhancement module, Improved anchor matching, and Progressive anchor loss~\cite{li2018dsfd} would be able to be adapted to the proposed model without revising the proposed structure.
Also, we used the MobileFaceNet~\cite{chen2018mobilefacenets}, the face variant of the MobileNet-V2~\cite{sandler2018mobilenetv2}, to the S3FD model instead of VGG-16 to see the effectiveness of the proposed method compared to the case of using the lightweight backbone network.

\textbf{Variations:} 
We applied the proposed recurrent scheme mainly into the FPN-based structure.
For the model, we designed three variations which have a different number of parameters, lighter one having $0.063$M parameters with $32$ channels for each feature maps, intermediate one having $0.1$M parameters with $48$ channels, and the heavier one with $64$ channels and $0.16$M parameters when designed as FPN.
See Appendix~\ref{app:backbone} for the detailed configuration of the backbone networks for each case.

Also, we tested different activation functions: ReLU, PReLU, and Leaky-ReLU for each model.
The negative slope of the Leaky-ReLU is set to $0.25$, which is identical to the initial negative slope of the PReLU.
In the following section, we will term each variation by a combination of abbreviations; \textit{EXTD-model-channel-activation}. For example, the term \textit{EXTD-FPN-32-PReLU} denotes the proposed model combined with FPN, with feature channel width $32$ and with activation function PReLU.

As an ablation, we also applied the proposed recurrent backbone into SSD-like structure as well. 
The ablation was trained and tested with the same conditions to the FPN-based version and abbreviated as \textit{SSD}.
Same as FPN case, for example, the term \textit{EXTD-SSD-32-PReLU} denotes the proposed model combined with SSD, with feature channel width $32$ and with activation function PReLU.

\subsection{Performance Analysis}
\begin{figure}[t]
\begin{center}
\includegraphics[width=0.99\linewidth]{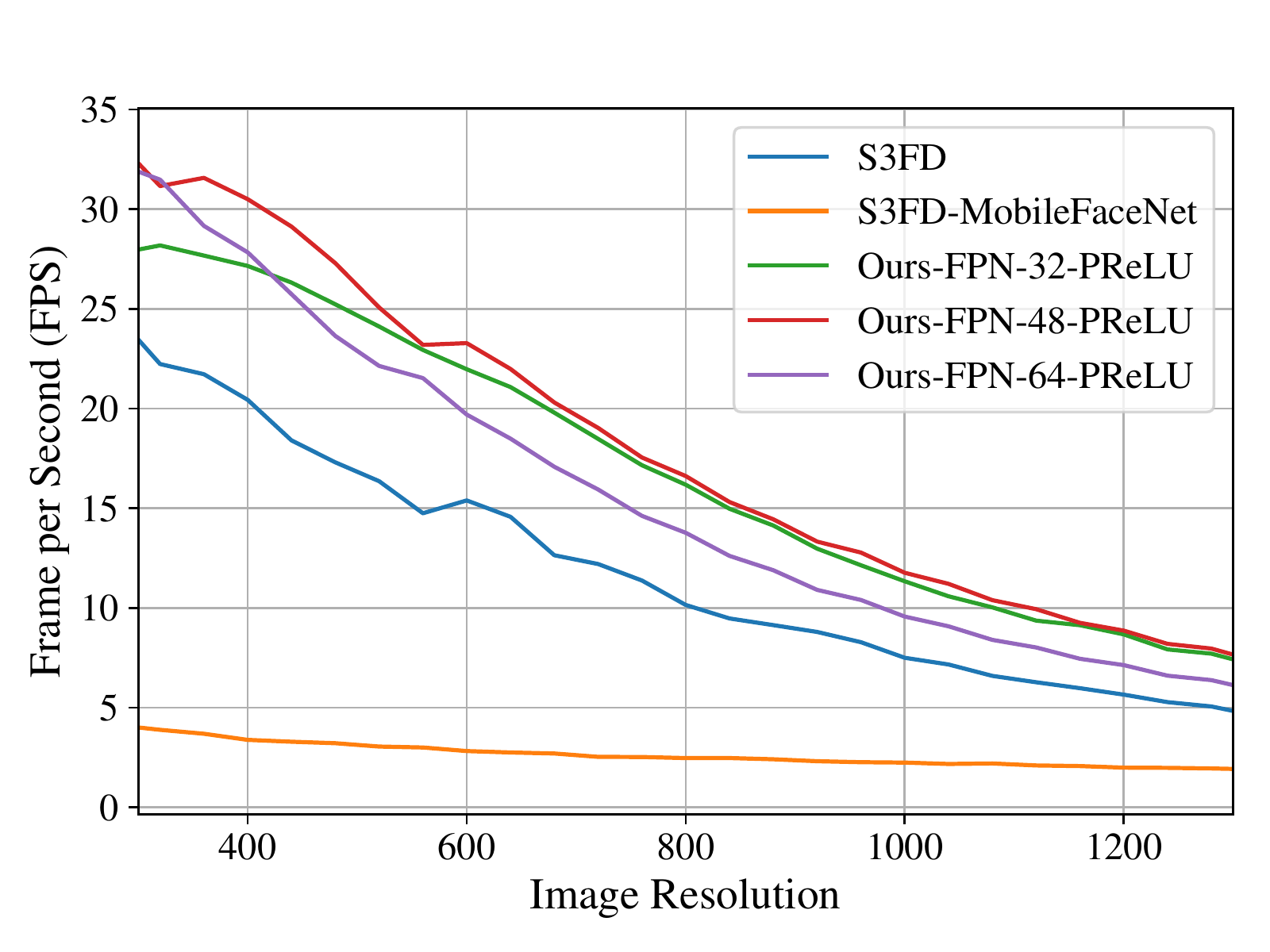}
\end{center}
\vspace{-2mm}
   \caption{Evaluation time given image resolutions (averaged 1000 trials each). The horizontal axis denotes the size of an image and the vertical axis shows the frame per second (FPS). The model with the higher value means that it has faster inference speed.}
\label{fig:fps}
\vspace{-2mm}
\end{figure}

In Table~\ref{table:wider_face_comp}, we list the quantitative evaluation results of face detection in WIDER FACE dataset and the comparison to the SOTA face detectors.
The table shows the mAP of the models on Easy, Medium, Hard cases for both validation and test sets of the dataset.
Also, the table includes model information such as their backbone networks, number of parameters, and total number of adder arithmetics (Madds). 
In Figure~\ref{fig:roc_curve}, the precision recall curve for the proposed and the other methods are presented.
Figure~\ref{fig:qual} shows the examples of the face detection results from images with various conditions.
In Figure~\ref{fig:fps}, we evaluate the latency of the models in terms of the resolution of images, which measured via a machine with CPU i7 core and NVIDIA TITAN-X. 
For a fair comparison, all the inference processes of the models are implemented by PyTorch 1.0.

\textbf{Comparison to the Existing Methods:}
The results in Table~\ref{table:wider_face_comp} shows that some variations of the proposed method achieved comparable performance to the baseline model S3FD.
Among lighter models and intermediate models, \textit{EXTD-FPN-32-PReLU} and \textit{EXTD-FPN-48-PReLU} each got a mAP score $3.4\%$ and $1.2\%$ lower than S3FD in WIDER Face hard validation set.
When compared to S3FD trained scratch, \textit{EXTD-FPN-64-PReLU} achieved even performances.
For the heavier version, we found that our FPN variant achieved nearly the same accuracy, only $0.3\%$ in WIDER FACE hard validation set and $0.8\%$ in test set to S3FD in spite of the huge model size and memory usage gaps.
It is meaningful in that the proposed detectors: lighter, intermediate, and heavier versions, are about $343$, $220$, and $138$ times lighter in model size and are $28.3$, $19.2$, and $11$ times lighter in Madds.

When compared to SOTA face detectors such as PyramidBox~\cite{tang2018pyramidbox} and  DSFD~\cite{li2018dsfd}, our best model \textit{EXTD-FPN-64-PReLU} achieved lower results.
The margin between PyramidBox and the proposed model on WIDER FACE hard case was $3.4\%$. 
Considering that PyramidBox inherits from S3FD and our model follows the equivalent training and detection setting to S3FD, our model would have a possibility to further increase the detection performance by adding the schemes proposed in PyramidBox.
The mAP gap to DSFD, which is tremendously heavier, is about $5.0\%$, but it would be safe to suggest that the proposed method offers more decent trade-off in that DSFD uses about 2860 times more parameters than the proposed method.
This is also meaningful result in that our method did not use any kind of pre-training of the backbone network using the other dataset such as ImageNet~\cite{imagenet_cvpr09}.
Figure~\ref{fig:roc_curve} shows the ROC curves of the proposed \textit{EXTD-FPN-64-PReLU} and the other methods. From the graphs, we can see that our method is included in the SOTA group of the detectors using heavyweight pre-trained backbone networks.

When it comes to our SSD-based variations, they got lower mAP results than FPN-based variants.
However, when compared with the S3FD version trained with MobileFaceNet backbone network, the proposed SSD variants achieved comparable or better detection performance.
It is a meaningful result in that the proposed variations have smaller feature map width, S3FD-MobileFaceNet holds feature map size of $[64, 128, 128, 128, 128, 128]$, and use the smaller number of layer blocks; inverted residual blocks same as MobileFaceNet, repeatedly. 
This shows that the proposed itertative scheme efficiently reduces the number of parameters without loss of accuracy.

Also, from the graph in Figure~\ref{fig:fps}, we showed that our EXTD achieved faster inference speed to the S3FD, which is considered as real-time face detector, in a wide range of an input image resolution.
This shows that the proposed face detector can safely alter S3FD without losing accuracy and with consuming much smaller capacity, as well as maintaining the inference speed.
It is interesting to note that the inference was much slow when using MobileFaceNet instead of  VGG-16. It would mainly be due to that MobileFaceNet version should pass more filters (48) than VGG-16 version (24), and the inference times of the filters including pooling, depth-wise, point-wise and ordinary convolutional filters are not that different in Pytorch implementation.

\textbf{Detection performance regarding the Face Scale: }
One notable characteristic of the proposed method captured from the evaluation is that our detector obtained better performance when dealing with a small size of faces.
From the table, we can see that our method achieved higher performance in WIDER FACE hard dataset than other cases.
Since the Easy and Medium cases are subsets of the Hard dataset, this means that the proposed method is especially fitted to capture small sized faces.
This tendency is commonly observed for different variations, for the different model architecture, and for the different channel widths.
This supports the proposition suggested in Section~\ref{sec:RFMap} that the proposed recurrent structure strengthens the feature map, especially for the lower-level feature maps, and hence enhance the detection performance of the small faces. 

\begin{table*}[]
\small
\centering
\tabcolsep=0.4cm
\begin{tabular}{@{}lccccc@{}}
\toprule
\multirow{2}{*}{Model} & 
\multirow{2}{*}{\# Params} & 
\multirow{2}{*}{\# Madds (G)} &
\multirow{2}{*}{\begin{tabular}[c]{@{}c@{}} \\ Easy (mAP)\end{tabular}} &
\multirow{2}{*}{\begin{tabular}[c]{@{}c@{}}WIDER FACE \\ Medium (mAP) \end{tabular}} &
\multirow{2}{*}{\begin{tabular}[c]{@{}c@{}} \\ Hard (mAP) \end{tabular}} \\
                                  &
                                  &                            
                                  &
                                  &                                  
                                  &        \\ \midrule

EXTD-SSD-32-ReLU                  & 0.056 M & 4.35 & 0.791 (-0.105)   & 0.770 (-0.115) & 0.629 (-0.196)\\
EXTD-SSD-32-LReLU                 & 0.056 M & 4.35 & 0.851 (-0.045)   & 0.836 (-0.049) & 0.736 (-0.089) \\
EXTD-SSD-32-PReLU                 & 0.056 M & 4.35 & 0.870 (-0.026)    & 0.855 (-0.030) & 0.757 (-0.068) \\ 
EXTD-FPN-32-ReLU                  & 0.063 M & 4.52 & 0.741 (-0.155)   & 0.735 (-0.150) & 0.642 (-0.182)\\
EXTD-FPN-32-LReLU                 & 0.063 M & 4.52 & 0.892 (-0.004)   & 0.884 (-0.001) & 0.824 (-0.001)\\
\textbf{EXTD-FPN-32-PReLU}        & 0.063 M & 4.52 & \textbf{0.896}   & \textbf{0.885}    & \textbf{0.825}\\ 
\midrule
EXTD-SSD-48-ReLU                  & 0.086 M   & 6.63  & 0.868 (-0.045)    & 0.852 (-0.052)   & 0.742 (-0.105) \\
EXTD-SSD-48-LReLU                 & 0.086 M   & 6.63  & 0.879 (-0.034)    & 0.860 (-0.044)   & 0.744 (-0.103) \\
EXTD-SSD-48-PReLU                 & 0.086 M   & 6.63  & 0.897 (-0.016)    & 0.879 (-0.025)   & 0.774 (-0.073) \\
EXTD-FPN-48-ReLU                  & 0.10 M   & 6.67  & 0.894 (-0.019)     & 0.885 (-0.019)     & 0.825 (-0.022) \\
EXTD-FPN-48-LReLU                 & 0.10 M   & 6.67  & 0.911 (-0.002)     & 0.901 (-0.003)     & 0.846 (-0.001) \\
\textbf{EXTD-FPN-48-PReLU}        & 0.10 M   & 6.67  & \textbf{0.913}    & \textbf{0.904}    & \textbf{0.847}\\
\midrule
EXTD-SSD-64-ReLU                  & 0.14 M   & 10.6  & 0.887 (-0.034)   & 0.867 (-0.044)  & 0.752 (-0.104)\\
EXTD-SSD-64-LReLU                 & 0.14 M   & 10.6  & 0.896 (-0.025)   & 0.878 (-0.033) & 0.769 (-0.087)\\
EXTD-SSD-64-PReLU                 & 0.14 M   & 10.6  & 0.905 (-0.016)   & 0.888 (-0.023) & 0.784 (-0.072) \\
EXTD-FPN-64-ReLU                  & 0.16 M   & 11.2  & 0.910 (-0.011)     & 0.900 (-0.011)     & 0.844 (-0.012)\\
EXTD-FPN-64-LReLU                 & 0.16 M   & 11.2  & 0.914 (-0.007)     & 0.906 (-0.005)     & 0.850 (-0.006)\\
\textbf{EXTD-FPN-64-PReLU}        & 0.16 M   & 11.2  & \textbf{0.921}    & \textbf{0.911}    & \textbf{0.856}\\
\midrule
\end{tabular}
\caption{Variation study on WIDER FACE validation dataset. 
The models with boldface denotes the representative models for each block. 
The value in the parentheses shows the margin between the best model in the block (written in boldface).
}
\vspace{-2mm}
\label{table:wider_face_ablation}
\end{table*}

\subsection{Variation Analysis}
The evaluation on the variations of the proposed EXTD is summarized in Table~\ref{table:wider_face_ablation}.
The table mainly consists of three blocks in rows. Each first, second, and third block lists the evaluation results from the smaller version (32 channels), intermediate version (48 channel), and the heavier version (64 channel) with applying different activation functions.

\textbf{Effect of the Model Architecture: }
From the table, we can find two common observations among the proposed variations. 
First, for all the different channel width, FPN based architecture achieved better detection performance compared to SSD based architecture, especially for detecting small faces.
The idea of expanding the number of layers for reaching the largest sized feature-map, for detecting the smallest size of objects, is a common strategy for SSD variant methods.
This approach assumes that typical SSD structure passes too small number of layers and hence, the resultant feature-map could not import much information useful for the detection task.
In the face detection task, this assumption seems to be correct in that the FPN based models notably achieved superior detection performance on small faces compared to SSD based models for all the cases.

Second, for both SSD based and FPN based model, channel width was another key factor for performance enhancement. 
As the channel width increased by $32$ to $64$, we can see that the detection accuracy significantly enhanced for all the cases; Easy, Medium, and Hard.
Considering that we used a smaller number of layers for $48$ and $64$ channel cases than the case with $32$ channel, this shows that having enough size of channel width is critical for embedding sufficient information to the feature map for detecting faces.

\textbf{Effect of the Activation functions: }
From the evaluation, we found that the choice of the activation function is another factor governing the detection performance of the proposed method. 
In all the cases including FPN based and SSD based structures, PReLU was the most effective choice when it comes to mAP, but the gap between Leaky-ReLU was not that significant for the FPN variants. 
When tested with SSD based architecture, PReLU outperformed Leaky-ReLU with larger margin than those using FPN structure.

It is worth noting that ReLU occurred notable performance decreases especially when the channel width was small for both SSD and FPN cases.
When the channel width was set to $32$, mAP for all the three cases were lower than $10\%$ to $20\%$ compared to those using other activation functions.
The decreases were alleviated as the channel width increased. 
When the channel width was $48$, the gap was about $2.2\%$, and in the channel width $64$ case, the margin was about $1.2\%$.
From the results, we conjecture that the nature of ReLU that set all the negative values to zero occurs information loss in the proposed iterative process since it makes the feature map too sparse, and this information loss would be much critical when the channel width is small.  
\section{Conclusion}
\label{sec:conclusion}
In this paper, we proposed a new face detector which significantly reduces the model sizes as well as maintaining the detection accuracy.
By re-using backbone network layers recurrently, we reduced the vast amount of the network parameters and also obtained comparable performance to recent deep face detection methods using heavy backbone networks.
We showed that our methods achieved very close mAP to the baseline S3FD only with hundreds time smaller parameters and with using tens time smaller Madd without using pre-training.
We expect that our method can be further improved by applying recent techniques of the SOTA detectors which integrated to S3FD.
\section*{Acknowledgement}
We are grateful to Clova AI members with valuable discussions, and to Jung-Woo Ha for proofreading the manuscript.

{\small
\bibliographystyle{ieee}
\bibliography{egbib}
}

\clearpage
\onecolumn
\begin{appendices}

\begin{figure*}[t]
\begin{center}
\includegraphics[width=0.95\linewidth]{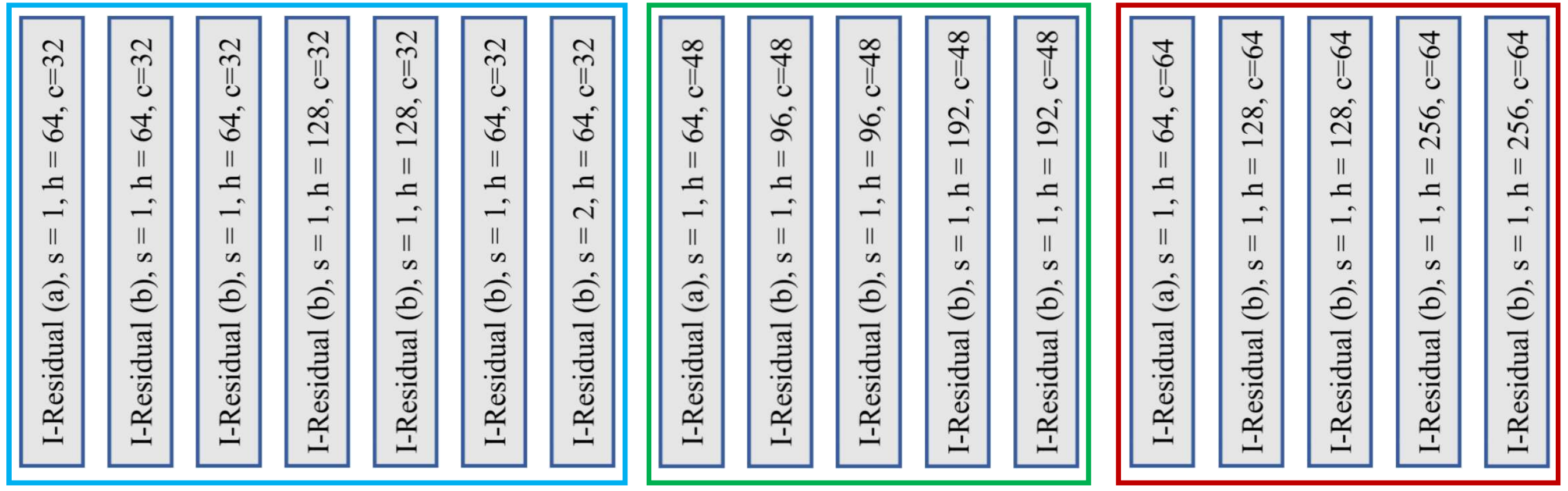}
\end{center}
   \caption{Backbone architectures for the recursive feature generation.}
\label{fig:archs}
\end{figure*}

\begin{table*}[]
\small
\centering
\tabcolsep=0.15cm
\begin{tabular}{ccccccccccccccc}
\toprule
Description & 
1 & 2& 3& 4& 5& 6& 7& 8& 9& 10& 11& 12& 13& 14\\ \midrule

I-Residual type     & (a) & (b) & (b) & (b) & (b) & (b) & (b) & (b) & (b) & (b) & (b) & (b) & (b) & (b) \\ 
Output channel width& 64 & 64 & 64 & 64 & 64 & 64 & 128 & 128 & 128 & 128 & 128 & 128 & 128 & 128 \\
Hidden channel width& 64 & 128 & 128 & 128 & 128 & 128 & 256 & 256 & 256 & 256 & 256 & 256 & 256 & 512 \\
Stride              & 2 & 1 & 1 & 1 & 1 & 2 & 2 & 1 & 1 & 1 & 1 & 1 & 1 & 2 \\

\midrule
\end{tabular}
\caption{Structure of MobileFaceNet backbone attached in S3FD. Three extra layers are attached to further reduce the feature map size.}

\label{table:s3fd_mobileFaceNet}
\end{table*}
\section{Implementation detail}
\label{app:implementation}
For training the proposed architecture, a stochastic gradient descent optimizer (SGD) with learning rate $1e^{-3}$, with $0.9$ momentum, $0.0005$ weight decay, and batch size $16$ is used. 
The training is conducted from scratch, and the network weights were initialized with He-method~\cite{he2015delving}.
The maximum iteration number is basically set to $240$K, and we drop the learning rate to $1e^{-4}$ and $1e^{-5}$ at $120$K and $180$K iterations.
Also, we test the architecture with twice larger iterations $480$K as well.  
In this case, the learning rate is dropped at $240$K and $360$K iterations.
Similar to the other networks using depth-wise separable networks~\cite{sandler2018mobilenetv2,li2018tiny}, further performance improvements were observed when training the network with larger iteration.

\section{Detailed Architecture Information}
\label{app:backbone}
Figure~\ref{fig:archs} shows the detailed structures of the backbone network for the variation having channel sizes $32$, $48$, and $64$.
The layers in `blue', `green', and `red' boxes in the figure each denotes the version of the proposed detectors having channel width to 32, 48, and 64. Each model has parameter size $0.063$M, $0.10$M, and $0.16$M respectively, when designed as FPN structure. The term `I-Residual' denotes the inverted residual block (a) and (b), where the configuration of the block is introduced in Figure~\ref{fig:backbone} of the paper.
The heavier versions which have $0.10$M, and $0.16$M model parameters are designed to have less number of parameters to reduce the parameter when compared to the lightest version. The results in the paper show that the width of the channels for each layer is more critical than the depth of the layers for the detection performance in the proposed model.

\section{Implementation of S3FD with MobileFaceNet Backbone} 
In the paper, we implemented the S3FD variation where the backbone network was set to MobileFaceNet instead of VGG-16. 
The backbone network consists of $14$ inverted residual blocks followed by 3x3 convolutional filter which has output channel width $64$ and stride two. 
The lowest-level inverted residual block is defined as in I-Residual (a), and the others are defined as I-Residual (b). 
The detailed setting of each blocks are described in Table~\ref{table:s3fd_mobileFaceNet}.
We added a classification and regression head at the bottom of layers 6, 7, and 14. 
After layer 14, three extra layers defined by 3x3 convolutional filter with output channel width 128 are attached. 
This extra layer setting is equivalent to original S3FD, and the resolutions of the feature maps are [64, 128, 128, 128, 128, 128] with total parameter number 1.2 million. 
The MobileFaceNet backbone itself is a reduced version of MobileNet-V2, and we only used the part of the MobileFaceNet layers. However, we can still see that the backbone network requires a large number of parameters which makes challenging to be embedded in smaller devices.

\end{appendices}

\end{document}